\documentclass{article}

\usepackage[preprint]{corl_2026} 

\usepackage{graphics} 
\usepackage{graphicx}
\usepackage{caption}
\usepackage{subcaption}
\usepackage{pgfplots}
\usepackage{pgfplotstable}
\pgfplotsset{compat=1.16}
\usepackage{amsmath} 
\usepackage{amssymb}  
\usepackage{enumitem} 
\usepackage{algorithm}
\usepackage{algpseudocode}

\title{TAP-VLA: Tactile Annotation Prompting for Vision Language Action Models}

\author{
  Mark Van der Merwe${^1}{^*}$, Mohamad Louai Shehab${^1}{^*}$, Jayjun Lee$^1$, Youngsun Wi$^1$\\ \textbf{Yinpei Dai$^2$, Dmitry Berenson$^1$, Nima Fazeli$^{1,2}$}\\
  $^1$Robotics Department, University of Michigan\\
  $^2$Computer Science and Engineering Department, University of Michigan\\
  $^*$Equal Contribution\\
  \texttt{\{markvdm,mlshehab,dmitryb,nfz\}@umich.edu}
}

\begin{document}
\maketitle


\begin{abstract}
Vision-Language-Action (VLA) models demonstrate impressive reasoning over visual, semantic, and spatial task variations by leveraging large-scale vision and language pre-training. They remain, however, largely blind to contact forces, which seldom manifest clearly in visual feedback but are central to contact-rich manipulation. Tactile sensing measures these forces directly, but integrating it into VLAs is difficult: tactile data is absent from the large-scale corpora used to pre-train VLAs, so adding it as a new input modality induces a distribution shift that erodes the very pre-training that makes VLAs effective. We propose Tactile Annotation Prompting for Vision-Language-Action models (TAP-VLA), a simple framework that supplies tactile feedback through visual augmentation rather than architectural change. TAP-VLA extracts shear fields from visuo-tactile sensors and overlays them as spatially-grounded vectors onto the multi-view RGB images the policy already consumes, yielding a clear, interpretable tactile cue in the VLA's native observation space. Because the architecture is untouched, the approach requires no tactile pre-training, adds negligible compute, and stays close to the pre-training distribution. Across four contact-rich tasks, TAP-VLA succeeds on 78\% of trials, compared to under 50\% for vision-only fine-tuning and alternative tactile-fusion baselines---including tasks where the baselines perform no better than chance. Video results can be found at \href{https://tap-vla.github.io/}{https://tap-vla.github.io/}.

\end{abstract}



\section{Introduction}

Vision-Language-Action (VLA) models offer a promising path toward general-purpose manipulation policies~\cite{ghosh2024octo,brohan2022rt}. By scaling imitation learning~\cite{zhao2023learning,chi2025diffusion} over large robot demonstration datasets~\cite{khazatsky2024droid,o2024open} and building on pre-trained vision-language backbones~\cite{black2025pi_,zitkovich2023rt,kim2025openvla,intelligence2025pi}, they generalize across diverse tasks and exploit the semantic reasoning of their backbones to drive open-world behavior~\cite{zitkovich2023rt,dai2026robomme}. Yet this generalization has been demonstrated almost entirely along \textit{visual}, \textit{semantic}, and \textit{spatial} axes.


\begin{figure*}
    \centering
    \includegraphics[width=\linewidth]{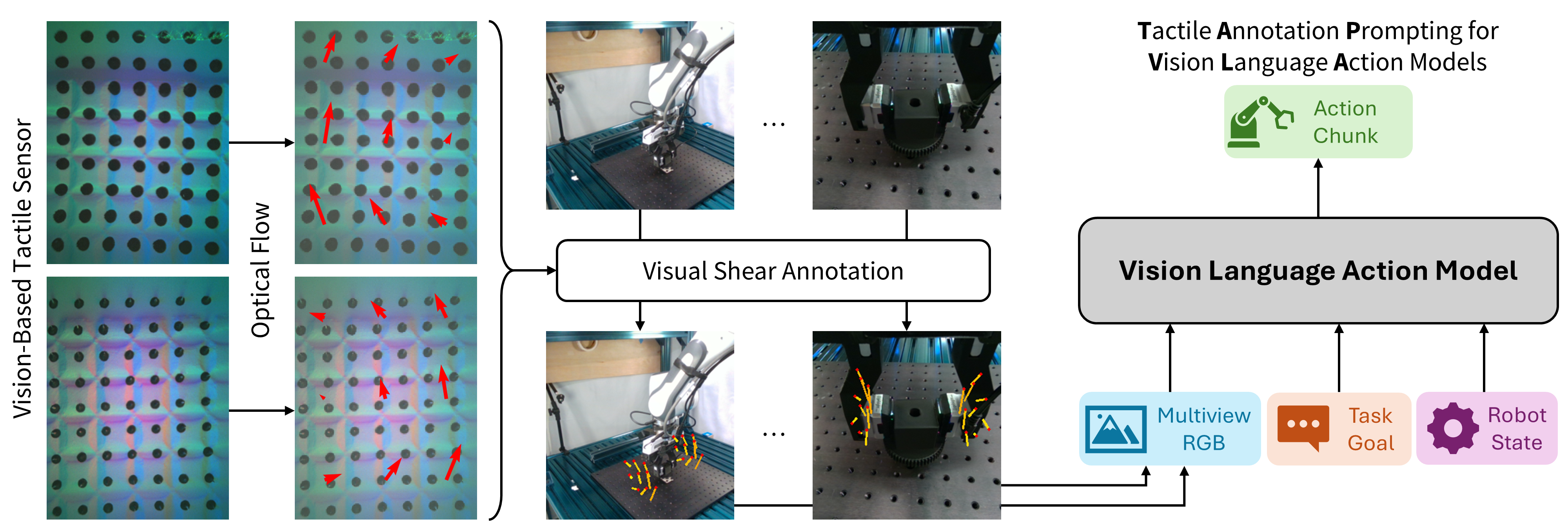}
    \caption{We propose Tactile Annotation Prompting for Vision-Language-Action models (TAP-VLA), bridging large-scale pre-trained behaviors with tactile feedback. We extract shear fields from the tactile image and annotate them directly into the multi-view RGB images during contact-rich tasks, such as this peg insertion. The annotated images are then provided alongside the language goal and robot state to fine-tune a pre-trained VLA for tactile-aware behavior.}
    \label{fig:intro_figure}
\end{figure*}

Many manipulation tasks, however, hinge on \textit{physical} quantities that these axes do not capture. Tool use requires regulating the forces a tool applies to its environment~\cite{8967889,oller2023manipulation}, and dynamic skills such as in-hand reorientation depend on object mass and friction~\cite{9341006}. Crucially, these forces seldom manifest in visual feedback: a camera captures geometry and appearance, but not the loads transmitted through contact. A vision-only VLA is therefore structurally blind to precisely the information that contact-rich manipulation depends on.

Tactile sensing offers a direct measurement of these forces, and visuomotor policies have exploited it both in isolation~\cite{bogert2025gromp,wu2025tacdiffusion} and alongside vision~\cite{hou2025adaptive,chen2025dexforce,pmlr-v205-chen23d}. Extending this to VLAs, however, fights what makes them effective. Their strength comes from large-scale pre-training, yet that data contains no tactile signals~\cite{khazatsky2024droid,o2024open}, and tactile data is itself too heterogeneous and scarce to pre-train at a comparable scale: force/torque sensors output 6D wrenches~\cite{choi2025coinft}, visuo-tactile sensors such as the GelSight output RGB images~\cite{yuan2017gelsight}, and existing datasets remain small and sensor-specific~\cite{fu2024touch,yang2022touch}. Tactile must thus be grafted on during fine-tuning, and prior methods do so by adding dedicated encoders or input branches~\cite{yu2025forcevla,zhang2025elucidating,cheng2025omnivtla}, inducing a distribution shift away from the pre-training the policy relies on. This raises our central question: \textit{can we endow a pre-trained VLA with tactile feedback without architectural changes or significant distribution shift?}

To address this question, we propose Tactile Annotation Prompting for Vision-Language-Action models (TAP-VLA), a lightweight visual augmentation that introduces no new inputs. Rather than routing tactile through a separate stream, we render it directly into the images the policy already consumes (Fig.~\ref{fig:intro_figure}). Inspired by the ability of VLMs to reason over image-level annotations~\cite{Shtedritski_2023_ICCV}, we extract shear fields from the GelSight visuo-tactile sensor~\cite{yuan2017gelsight} and draw down-sampled, spatially-grounded shear vectors into the multi-view RGB observations. The result is a simple, interpretable tactile cue that lives in the VLA's native observation space: it leaves the architecture untouched, requires no tactile pre-training, and adds negligible compute.

We evaluate TAP-VLA on four real-world manipulation tasks that demand reasoning about object mass, center of mass, and contact alignment. TAP-VLA succeeds on 78\% of trials versus under 50\% for every baseline, and is the only method to perform reliably across all four tasks---including settings where the alternatives are no better than chance. In summary, we contribute:
\begin{itemize}[leftmargin=1.4em, itemsep=2pt, topsep=1pt, parsep=0pt]
    \item \textbf{Tactile annotation prompting}, a framework that endows pre-trained VLAs with tactile feedback by rendering it as visual annotations.
    \item \textbf{A shear-field annotation pipeline} that converts raw visuo-tactile images into spatially-grounded shear vectors and projects them into the multi-view RGB observations of the policy.
    \item \textbf{Real-world evaluation} on four contact-rich tasks, benchmarking against raw-tactile fusion, a learned tactile encoder, and vision-only VLA baselines.
\end{itemize}

\vspace{-9pt}
\section{Related Work}
\vspace{-6pt}

\textbf{Tactile VLA Models:} Several recent works incorporate force and tactile sensing into VLAs. Hao et al.~\cite{hao2025tla} train a Tactile-Language-Action model from scratch on Qwen2-VL~\cite{wang2024qwen2} with tactile \textit{replacing} vision; Zhang et al.~\cite{zhang2025vtla} add vision back in to form a Vision-Tactile-Language-Action model on the same backbone. Training from scratch, however, forgoes the policy pre-training that drives VLA success and demands large-scale tactile data. Other works instead start from pre-trained VLAs such as $\pi_0$~\cite{BlackK-RSS-25} or Octo~\cite{ghosh2024octo} and add dedicated encoders for force~\cite{yu2025forcevla}, joint torque~\cite{zhang2025elucidating}, taxel~\cite{huang2025tactile}, or vision-based tactile~\cite{cheng2025omnivtla,beyond2025jones} inputs, fine-tuning on newly-collected demonstrations. Several augment this with auxiliary objectives to improve cross-modal reasoning---cross-modal contrastive losses~\cite{cheng2025omnivtla,beyond2025jones}, cross-modal language generation~\cite{beyond2025jones}, or torque prediction~\cite{zhang2025elucidating}. These approaches all rely on architectural changes that induce a distribution shift from the original pre-training, which can inhibit effective policy learning.

\noindent\textbf{Visual Augmentation in VLA Models:} Overlaying visual annotations onto input images can substantially enhance the reasoning and grounding of VLMs and VLAs, spanning general-purpose visual prompting to robot-specific action representations rendered directly in image space.

\textit{Visual prompting for VLMs.}
Pixel-space marks can steer VLMs without changing model weights.
Shtedritski et al.~\cite{Shtedritski_2023_ICCV} drew a red circle around a region to direct CLIP's attention, achieving strong zero-shot referring expression comprehension.
Set-of-Mark prompting~\cite{yang2023set} overlays alphanumeric labels on segmented regions, unlocking fine-grained grounding in GPT-4V.
Li et al.~\cite{li2024visual} generalize this with a universal framework supporting strokes, boxes, and points as visual prompts.

\textit{Visual prompting for robotic reasoning.}
The same idea has been adapted to elicit spatially-grounded robot actions.
PIVOT~\cite{nasiriany2024pivot} overlays candidate action proposals and iteratively queries a VLM to refine them, enabling zero-shot control.
RoboPoint~\cite{robopoint} instruction-tunes a VLM on synthetic data to predict 2D keypoint affordances from language, which are projected into 3D for execution.
RT-Affordance~\cite{rt-affordance} renders predicted end-effector affordances as an intermediate representation, conditioning a downstream policy on the augmented observations.

\textit{Trajectory and flow overlays as action representations.}
Another family encodes motion as visual overlays.
RT-Trajectory~\cite{rt-trajectory} conditions policies on color-graded 2D trajectory sketches of the end-effector.
Im2Flow2Act~\cite{flow-as-interface} uses predicted object-centric optical flow as a cross-domain interface, enabling sim-to-real transfer specified through human videos.
TraceVLA~\cite{trace-vla} overlays dense point trajectories onto observations to enhance spatial-temporal awareness over OpenVLA.
GeniMa~\cite{genima} instead \emph{generates} images with target joint positions rendered as colored spheres.
AimBot~\cite{aimbot} projects simple shooting lines and scope reticles derived from end-effector pose and depth, with negligible compute overhead.

These methods augment images with geometric, kinematic, or motion cues, but none convey contact-level forces. TAP-VLA fills this gap by rendering tactile shear fields directly onto image observations, given the VLA explicit force information that is otherwise absent from visual inputs alone.
\vspace{-7pt}
\section{Method}
\vspace{-5pt}

TAP-VLA integrates tactile feedback into a pre-trained VLA by rendering it as visual annotations on the policy's existing camera inputs. This preserves the input signature of the base VLA and avoids the distribution shift incurred by adding new input streams at fine-tuning time. We first formalize the policy (\S\ref{sec:problem_statement}), then describe the annotation procedure that produces the augmented images (\S\ref{sec:annotation}).

\begin{figure*}[t]
    \centering
    
    \begin{subfigure}[b]{0.48\textwidth}
        \centering
        \includegraphics[width=\linewidth]{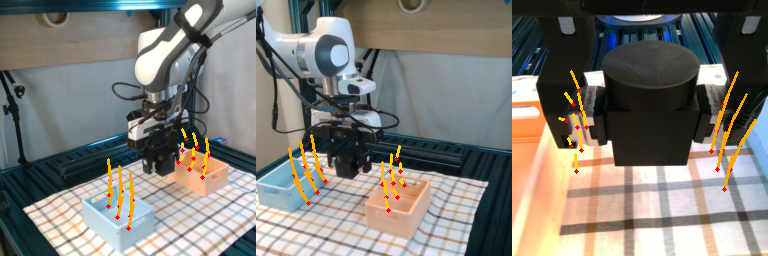}
        \caption{\texttt{medicine} (Full)}
        \label{fig:medicine_ann_1}
    \end{subfigure}
    \hfill
    \begin{subfigure}[b]{0.48\textwidth}
        \centering
        \includegraphics[width=\linewidth]{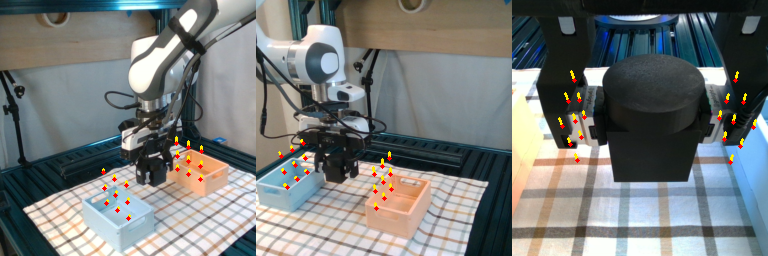}
        \caption{\texttt{medicine} (Empty)}
        \label{fig:medicine_ann_2}
    \end{subfigure}
    
    \vspace{0.5em}
    
    \begin{subfigure}[b]{0.48\textwidth}
        \centering
        \includegraphics[width=\linewidth]{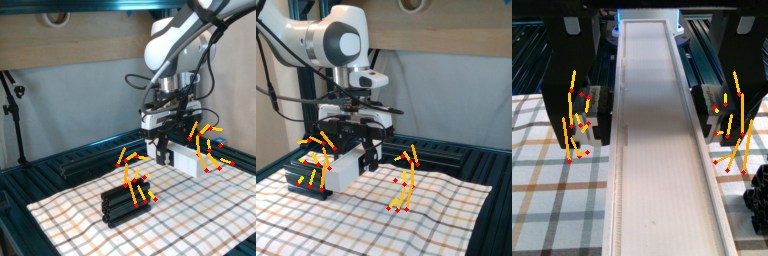}
        \caption{\texttt{balance}}
        \label{fig:balance_ann}
    \end{subfigure}
    \hfill
    \begin{subfigure}[b]{0.48\textwidth}
        \centering
        \includegraphics[width=\linewidth]{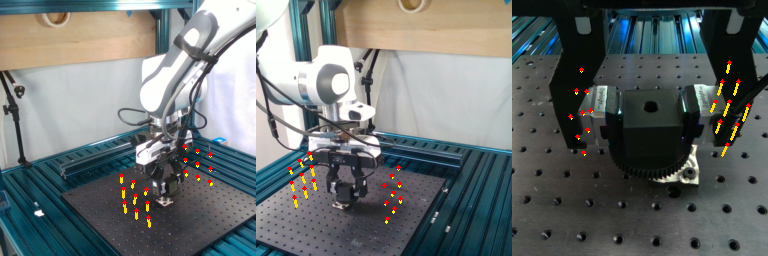}
        \caption{\texttt{gear}}
        \label{fig:gear_ann}
    \end{subfigure}
    
    \vspace{0.5em}
    \begin{subfigure}[b]{0.48\textwidth}
        \centering
        \includegraphics[width=\linewidth]{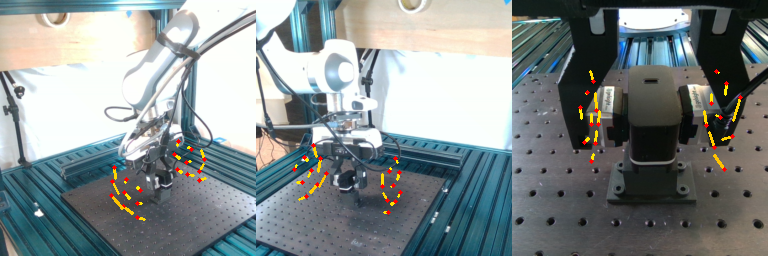}
        \caption{\texttt{plug}}
        \label{fig:plug_ann}
    \end{subfigure}

    \caption{Example shear-based visual annotations of multi-view RGB for our four tasks. The shear visually differentiates the full vs. empty medicine bottle (figs.~\ref{fig:medicine_ann_1} vs.~\ref{fig:medicine_ann_2}), elucidates the center of mass of the object to be balanced (fig.~\ref{fig:balance_ann}) and shows the misalignment of the gear or plug when being inserted (fig.~\ref{fig:gear_ann} and~\ref{fig:plug_ann}).}
    \label{fig:shear_annotation}
\end{figure*}

\subsection{Problem Statement}
\label{sec:problem_statement}

A VLA maps multi-view RGB images $\{I_t^i\}_{i=1}^K$ with $I_t^i\in\mathbb{R}^{H_C\times W_C\times 3}$, proprioception $\theta_t\in \mathbb{R}^N$, and a language instruction $l$ to an action chunk $a_{t:t+H}\in \mathbb{R}^{H\times M}$:
\begin{equation}
    \text{VLA}: \{I_t^1,\dots,I_t^K\}\,\times\,\theta_t\,\times\,l \;\to\; a_{t:t+H}.
\end{equation}
Our tactile setup uses a parallel-jaw gripper in which each finger carries a visuo-tactile sensor that outputs an RGB image, $I_t^L, I_t^R \in \mathbb{R}^{H_T\times W_T\times 3}$. The natural way to incorporate these signals is to widen the input signature with a new modality; we instead leave the signature untouched and route the tactile information through the images the policy already consumes.

Concretely, we apply a per-view annotation function $f$ that overlays the tactile shear onto each camera image, producing annotated images $\hat{I}_t^i = f(I_t^i, I_t^L, I_t^R, \theta_t, P^i)$ of the same dimension as $I_t^i$, where $P^i$ is the projection matrix of camera $i$. The fine-tuned policy then operates on the annotated images while preserving its original mapping:
\begin{equation}
    \text{TAP-VLA}: \{\hat{I}_t^1,\dots,\hat{I}_t^K\}\,\times\,\theta_t\,\times\,l \;\to\; a_{t:t+H}.
\end{equation}
The input/output signature is identical to the base VLA; only the pixel content of the multi-view images differs. Beyond the tactile sensors themselves, the only additional assumption we make is that the cameras are calibrated, so the projection matrices $\{P^i\}$ are known. The remainder of this section describes $f$.

\subsection{Visual Annotation}
\label{sec:annotation}

The visuo-tactile sensors we use yield raw RGB images $I_t^L, I_t^R$ from a marker-printed gel membrane that deforms under contact~\cite{yuan2017gelsight,taylor2022gelslim}. Recent work has shown that optical flow of these markers is a useful intermediate representation~\cite{van2026simultaneous,bogert2024built}; we adopt this \emph{shear field} as the tactile signal we annotate into each camera view. The procedure has three stages: (i) compute a shear field from each tactile image, (ii) lift the per-finger shear vectors into 3D in the fingertip frame, and (iii) project them into every camera view and draw them as colored lines. Algorithm~\ref{alg:annotation} summarizes the full pipeline; we walk through the stages below.

\begin{algorithm}[t]
\caption{TAP-VLA shear-field annotation $f$ (one timestep)}
\label{alg:annotation}
\begin{algorithmic}[1]
\Require Camera images $\{I_t^i\}_{i=1}^K$; tactile images $I_t^L, I_t^R$ and current base images $I_r^L, I_r^R$; fingertip poses $T_t^L, T_t^R$ (from $\theta_t$); camera projections $\{P^i\}_{i=1}^K$; scale $\alpha$.
\Statex
\State \textbf{Compute shear per finger.}
\For{$s \in \{L, R\}$}
    \State $S_t^s \gets \textsc{Farneback}(I_r^s,\, I_t^s)$
    \State $\tilde{S}_t^s \gets \textsc{MeanPool}(S_t^s) \in \mathbb{R}^{H_S\times W_S\times 2}$
\EndFor
\State \textbf{Lift to 3D in fingertip frame.}
\For{$s \in \{L, R\}$}
    \State $G_t^{s,\text{s}} \gets$ regular grid on the sensor surface
    \State $G_t^{s,\text{e}} \gets G_t^{s,\text{s}} + \alpha\,\tilde{S}_t^s$ \Comment{2D shear lifted to 3D with zero normal component}
\EndFor
\State \textbf{Project into each camera and draw.}
\For{$i = 1, \ldots, K$}
    \State $\hat{I}_t^i \gets I_t^i$
    \For{$s \in \{L, R\}$, each grid index $j$}
        \State $u_j^{\text{s}}, u_j^{\text{e}} \gets P^i\,T_t^s\,[\bar{G}_t^{s,\text{s}}[j];\ \bar{G}_t^{s,\text{e}}[j]]$
        \State Draw colored line $u_j^{\text{s}} \to u_j^{\text{e}}$ on $\hat{I}_t^i$ (color $\propto \lVert\tilde{S}_t^s[j]\rVert$)
    \EndFor
\EndFor
\State \Return $\{\hat{I}_t^i\}_{i=1}^K$
\end{algorithmic}
\end{algorithm}

\subsubsection{Tactile Shear Field Computation}

At each timestep, and for each finger, we compute the shear field using the Farneback algorithm~\cite{farneback2003two}, which measures the optical flow of the gel markers from a base image $I_r$ to the current image $I_t$:
\begin{equation}
    S_t = \text{Flow}(I_r, I_t), \qquad S_t \in \mathbb{R}^{H_T\times W_T\times 2}.
\end{equation}
$S_t$ captures the per-pixel deformation caused by forces acting on the sensor.

\noindent\textbf{Base-image reset.} Deformations on the sensor come from two sources: \emph{internal} forces produced by the grasp itself, and \emph{external} forces from gravity or extrinsic contacts. For the tasks we consider, the external component is the informative one---it reveals object mass, center of mass, and contact misalignment, while the internal squeeze is largely constant across attempts. We therefore \emph{reset} the base image $I_r$ per finger whenever its grasp state changes (grasped $\leftrightarrow$ released). Subsequent shear is then measured relative to the freshly-grasped state to isolate externally-induced deformations.

\subsubsection{Visual Augmentation with Shear Fields}

We describe the annotation for a single shear field; the procedure is repeated for each finger and each camera view.

\noindent\textbf{Downsample and lift to 3D.} We mean-pool the shear field to a lower-resolution grid $\tilde{S}_t \in \mathbb{R}^{H_S\times W_S\times 2}$. We then place a regular grid of 3D points $G_t^{\text{s}} \in \mathbb{R}^{H_S\times W_S\times 3}$ (denoting the \emph{start points}) on the sensor surface, expressed in the fingertip frame. Each pooled shear vector defines an \emph{end point} by displacing the corresponding start point in the plane of the sensor:
\begin{equation}
    G_t^{\text{e}} = G_t^{\text{s}} + \alpha\,\tilde{S}_t,
\end{equation}
where $\tilde{S}_t$ is lifted to 3D by appending a zero in the sensor-normal direction, and $\alpha$ scales the displacement (set per-task from the $95$th-percentile shear magnitude; see \S\ref{sec:exp}).

\noindent\textbf{Project and draw.} For each camera $i$, we transform the grid points from the fingertip frame to the world frame using the fingertip pose $T_t \in SE(3)$ (recovered from proprioception $\theta_t$), then project to pixels with the camera matrix $P^i$, using $\bar{\cdot}$ to denote homogeneous coordinates:
\begin{equation}
    \bar{u} = P^i\,T_t\,\bar{p}, \qquad p \in G_t^{\text{s}} \cup G_t^{\text{e}}.
\end{equation}
We then draw a colored line in image space from each projected start point to its corresponding end point, with color conveying shear magnitude. Repeating this for the left and right fingers across all camera views yields the annotated images $\{\hat{I}_t^1,\dots,\hat{I}_t^K\}$ used to fine-tune the pre-trained VLA. Example annotations are shown in Fig.~\ref{fig:shear_annotation}: the shear reveals task-relevant features such as object mass, center of mass, and contact misalignment.
\section{Experiments}\label{sec:exp}


\begin{figure}[t]
    \centering
    \begin{tikzpicture}
    \begin{axis}[
        ybar,
        width=\linewidth,
        height=0.25\textheight,
        bar width=11pt,
        ymin=0,
        ymax=1.05,
        xmin=-0.55,
        xmax=4.55,
        ylabel={Success Rate},
        xtick={0,1,2,3,4},
        xticklabels={medicine,balance,gear,plug,Overall},
        x tick label style={font=\small},
        ymajorgrids=true,
        grid style=dashed,
        legend style={
            at={(0.5,1.00)},
            anchor=south,
            draw=none,
            font=\small,
            legend columns=4,
            /tikz/every even column/.append style={column sep=0.25cm}
        },
        legend image code/.code={
            \draw[#1, draw=black!30] (0cm,-0.08cm) rectangle (0.13cm,0.13cm);
        },
        nodes near coords,
        nodes near coords align={vertical},
        every node near coord/.append style={
            font=\scriptsize,
            yshift=2pt,
            black
        },
        point meta=explicit symbolic,
        ylabel style={font=\small},
        tick label style={font=\small},
    ]

    \addplot+[
        fill=blue!60,
        draw=black!30
    ] coordinates {
        (0,0.3667) [11]
        (1,0.3667) [11]
        (2,0.4000) [12]
        (3,0.5333) [16]
        (4,0.4167) [50]
    };

    \addplot+[
        fill=orange!70,
        draw=black!30
    ] coordinates {
        (0,0.3333) [10]
        (1,0.4333) [13]
        (2,0.6333) [19]
        (3,0.5667) [17]
        (4,0.4917) [59]
    };

    \addplot+[
        fill=purple!55,
        draw=black!30
    ] coordinates {
        (0,0.4333) [13]
        (1,0.2000) [6]
        (2,0.4667) [14]
        (3,0.6000) [18]
        (4,0.4250) [51]
    };

    \addplot+[
        fill=green!65!black,
        draw=black!30
    ] coordinates {
        (0,0.8000) [24]
        (1,0.8333) [25]
        (2,0.7667) [23]
        (3,0.7333) [22]
        (4,0.7833) [94]
    };

    \legend{
        $\pi_{0.5}$,
        $\pi_{0.5}$ + tactile,
        $\pi_{0.5}$ + encoder,
        TAP-VLA (ours)
    }

    \draw[densely dashed, black!60]
        (axis cs:3.5,0) -- (axis cs:3.5,1.05);

    \end{axis}
    \end{tikzpicture}

    \caption{Quantitative performance across our tasks. Each method was run 30 times per task. Numbers above bars indicate successful runs.}
    \label{fig:quantitative_results}
\end{figure}
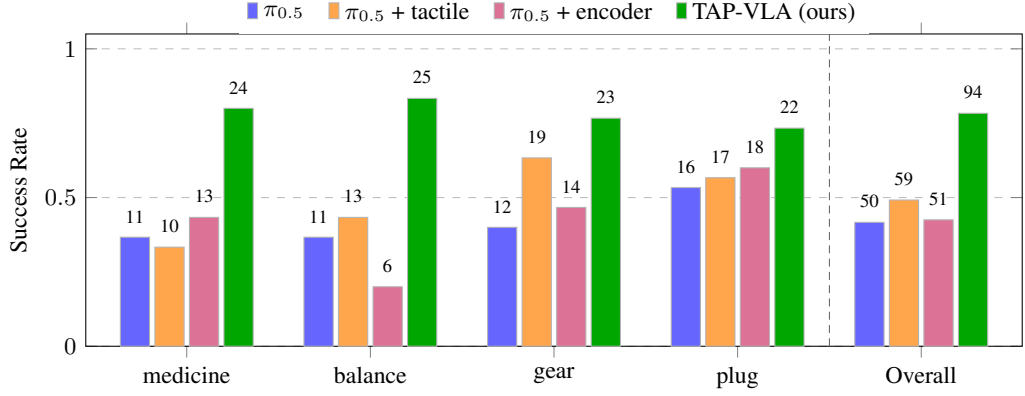

\noindent\textbf{Robot Setup.} We conduct our experiments on a 7-DoF Franka Emika Panda robot. We attach GelSight sensors~\cite{yuan2017gelsight} to the default Franka gripper. We utilize three RealSense cameras: two RealSense D435 cameras on the left and right hand side of the robot providing over-the-shoulder views and a wrist-mounted RealSense D405 providing a local view. We calibrate the external cameras to the robot and use the robot proprioception to track the wrist camera pose. Example views from each camera can be seen in Fig.~\ref{fig:shear_annotation}.

\noindent\textbf{Tasks.} We setup four tasks to test the ability of the fine-tuned VLA model to reason about physical forces during interaction.
\begin{enumerate}[leftmargin=1.4em, itemsep=2pt, topsep=1pt, parsep=0pt]
    \item \texttt{medicine} - the robot is presented with an opaque ``medicine'' bottle and two containers. The task is to place the bottle in the orange container if it is full or in the blue container if it is empty. This task involves reasoning about the mass of the object. During our experiment, the bottle weighs 38g when empty and 422g when full.
    \item \texttt{balance} - the robot is tasked with balancing one object on another in the scene. The object to be balanced has a fixed mass, but is designed so that the center of mass can be adjusted. The robot must reason about the center of mass to stably place the object without toppling. We run the task shifting the center of mass between three locations: centered and at each end of the object.
    \item \texttt{gear} - the robot starts grasping a gear object from the IndustReal benchmark~\cite{tang2023industreal} and must place the gear onto a peg mounted on the tabletop. This requires reasoning online about extrinsic forces during the interaction to align the gear and the peg.

    \item \texttt{plug} - the robot starts grasping a cordless plug end from the IndustReal benchmark~\cite{tang2023industreal} and should insert it into a socket mounted on the tabletop. Similar to \texttt{gear}, the robot should reason about the interaction forces to be able to align the prongs with the socket. 
\end{enumerate}
We collect demonstration data by teleoperating the robot via the Oculus Quest 2 VR device, recording observations and action labels at roughly 10Hz. We collect 100 demonstrations per task, varying spatial arrangements and lighting throughout.

\noindent\textbf{Implementation Details.} We implement our method utilizing the $\pi_{0.5}$ VLA model~\cite{black2025pi_}. We input the shear-annotated over-the-shoulder and wrist images, along with the robot joint state (7-dim) and the task language label (see Fig.~\ref{fig:intro_figure}). Actions are predicted as delta joint angles along with a binary gripper open/close command (8-dim). We use an action horizon $H=10$. We train with AdamW for 20k steps with 32 batch size.

To scale the shear annotations into the image in a reliable and interpretable way, we compute the 95th percentile of the shear vector magnitudes for each task during training and scale the vectors accordingly. We also clip vectors if they exceed twice the magnitude of the 95th percentile, to avoid spurious shears causing unexpected visuals.

\noindent\textbf{Baselines.} We compare TAP-VLA against three baselines that span the natural ways to incorporate tactile feedback into a VLA:
\begin{itemize}[leftmargin=1.4em, itemsep=2pt, topsep=1pt, parsep=0pt]
    \item $\pi_{0.5}$: we finetune the base VLA model \textit{only} on the visual feedback.
    \item $\pi_{0.5}$ + Tactile: we include the \textit{raw} tactile images as auxiliary visual inputs to the VLA model, treating them as additional ``viewpoints.''
    \item $\pi_{0.5}$ + Encoder: following recent works~\cite{zhang2026tacvla,beyond2025jones,cheng2025omnivtla} that decouple tactile sensing from the pretrained visual stream of a VLA, we feed the two raw gelsight images into a small from-scratch convolutional encoder and concatenate the resulting tokens into the $\pi_{0.5}$ prefix. Concretely, the tactile encoder is a four-layer stride-2 CNN with channel widths $(16, 32, 64, 128)$, $3\times3$ kernels, GELU activations, and per-layer LayerNorm. Each gelsight frame is resized to $64\times64$ before being fed to the stack, producing a $4\times4\times128$ feature map that is flattened into 16 patch tokens and projected to the LLM width (2048). The same encoder weights are shared between the left and right gelsights and are trained jointly alongside the rest of the VLA.
\end{itemize}
 All methods are trained under the same optimizer, learning rate schedule, dataset, and number of steps as TAP-VLA.

\subsection{Results}

\begin{figure*}[t]
    \centering
    
    \begin{subfigure}{\textwidth}
        \centering
        \includegraphics[width=\textwidth]{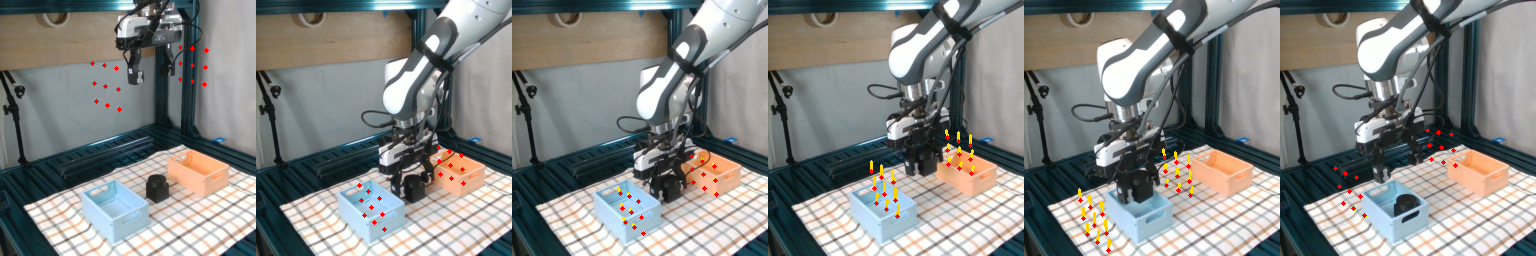}
        \caption{\texttt{medicine} (empty)}
        \label{fig:rollout_medicine_empty}
    \end{subfigure}
    
    \vspace{0.5em}
    
    \begin{subfigure}{\textwidth}
        \centering
        \includegraphics[width=\textwidth]{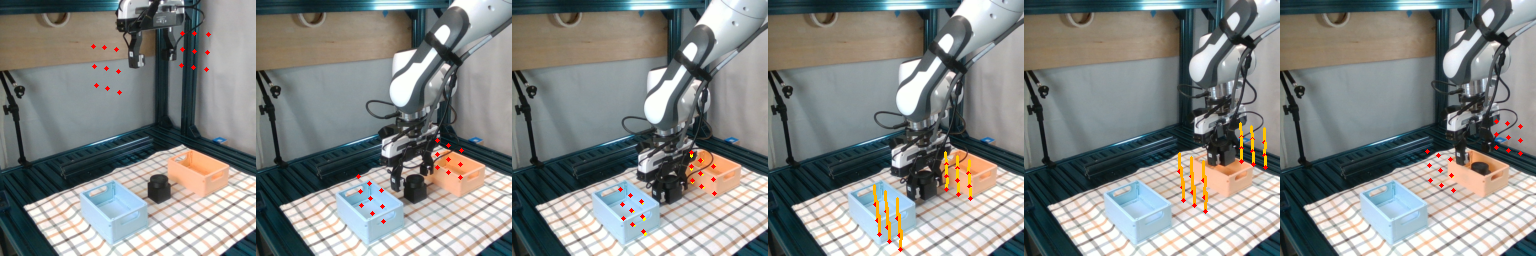}
        \caption{\texttt{medicine} (full)}
        \label{fig:rollout_medicine_full}
    \end{subfigure}
    
    \vspace{0.5em}
    
    \begin{subfigure}{\textwidth}
        \centering
        \includegraphics[width=\textwidth]{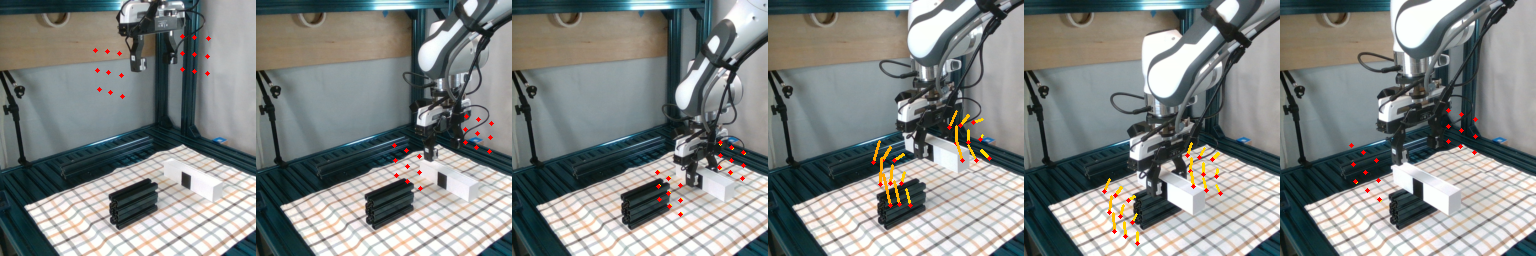}
        \caption{\texttt{balance} (CoM ``far'')}
        \label{fig:rollout_balance}
    \end{subfigure}

    \vspace{0.5em}
    
    \begin{subfigure}{\textwidth}
        \centering
        \includegraphics[width=\textwidth]{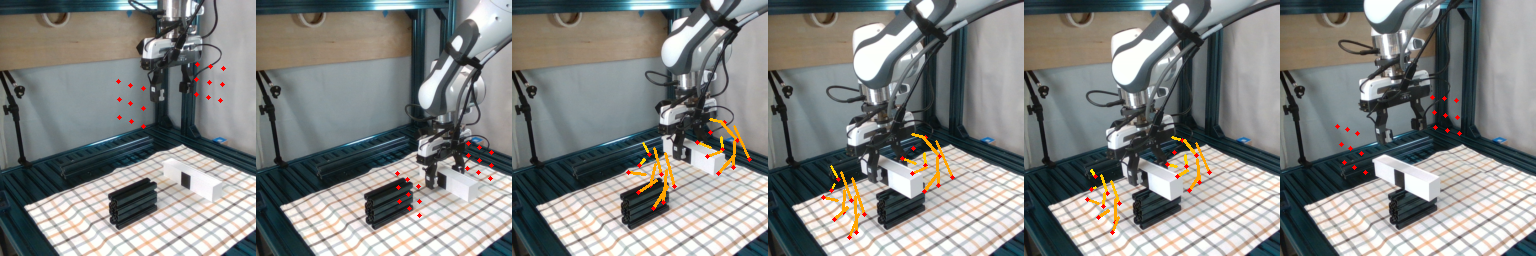}
        \caption{\texttt{balance} (CoM ``near'')}
        \label{fig:rollout_balance_2}
    \end{subfigure}
    
    \vspace{0.5em}
    
    \begin{subfigure}{\textwidth}
        \centering
        \includegraphics[width=\textwidth]{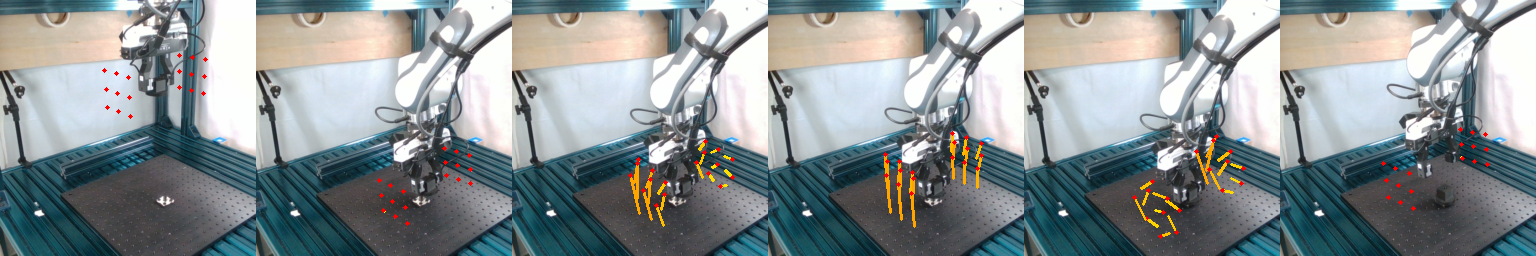}
        \caption{\texttt{gear}}
        \label{fig:rollout_gear}
    \end{subfigure}

    \begin{subfigure}{\textwidth}
        \centering
        \includegraphics[width=\textwidth]{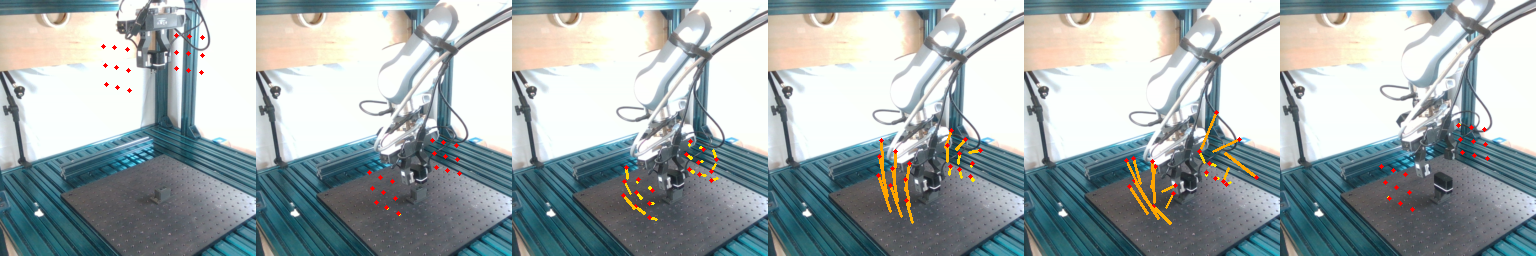}
        \caption{\texttt{plug}}
        \label{fig:rollout_plug}
    \end{subfigure}
    \caption{Sample successful rollouts of our proposed method, TAP-VLA, across our four test tasks.}
    \label{fig:rollout_qual}
\end{figure*}

Figure~\ref{fig:quantitative_results} reports success rates across 30 trials per task. TAP-VLA achieves the best overall performance (94/120 successes) and leads on every individual task. The three baselines yield comparable overall results despite their differing tactile-fusion strategies---adding tactile data alone is insufficient; \emph{how} it is injected into the policy is what dictates performance.

On \texttt{medicine} and \texttt{balance}, all three baselines perform at or below chance (50\% for \texttt{medicine}; $1/3$ for \texttt{balance}). None reliably identifies the relevant physical property (in particular, the bottle's mass or the object's center of mass) from the available signals. TAP-VLA, in contrast, succeeds on the majority of trials: 24/30 on \texttt{medicine} and 25/30 on \texttt{balance}.

On the \texttt{gear} and \texttt{plug} insertion tasks, TAP-VLA leads in success rate and exhibits robust error-correcting behavior after failed insertion attempts. The two tactile-fusion baselines also show some corrective behavior, while the vision-only $\pi_{0.5}$ frequently wedges or jams. This suggests the robot can reach coarse alignment with the peg but cannot complete the insertion.

Fig.~\ref{fig:rollout_qual} shows qualitative TAP-VLA rollouts on each task, with one annotated over-the-shoulder view per task. The annotations reveal whether the medicine bottle is full or empty, guiding placement into the correct bin (Figs.~\ref{fig:rollout_medicine_empty},~\ref{fig:rollout_medicine_full}); the implied center of mass, guiding the placement strategy on \texttt{balance} (Figs.~\ref{fig:rollout_balance},~\ref{fig:rollout_balance_2}); and the continuously-changing shears that drive corrective alignment during insertion on \texttt{gear} and \texttt{plug} (Figs.~\ref{fig:rollout_gear},~\ref{fig:rollout_plug}).

\section{Discussion}
We presented TAP-VLA, a lightweight scheme that renders tactile shear as visual annotations on the policy's existing camera views. Our method requires no architectural changes, adding minimal compute, and remaining close to the pre-training distribution. Across four contact-rich tasks, TAP-VLA outperformed both vision-only fine-tuning and two tactile-fusion baselines with access to the same tactile signal. The encoder-baseline result in particular suggests that, for a pre-trained VLA, how a new modality is presented may matter as much as the raw signal itself: tactile rendered into the model's native observation space appears easier to exploit than through a separate input stream.

The recipe may not be specific to GelSight or shear: in principle, any signal that admits a 2D rendering (e.g., other tactile representations, force/torque time-series, audio spectrograms, or depth overlays) could be supplied to a pre-trained VLA in the same way. Whether this low-friction transfer generalizes beyond visuo-tactile shear remains an open question worth pursuing.


\section{Limitations}

Several limitations remain. The shear field and subsequent downsampling discard tactile information that may matter for some tasks---most directly, the detailed grasp geometry conveyed by the raw GelSight image. The annotations can also occlude task-relevant visual content in cluttered scenes (mitigatable by additionally providing an unannotated view, at the cost of some redundancy), and a multi-fingered hand with many sensors would yield a denser overlay that likely reduces interpretability. We view TAP-VLA as complementary to longer-term efforts on large-scale tactile pre-training: a way to leverage existing pre-trained VLAs today, while data and architectures for native tactile reasoning continue to mature.

\clearpage


\bibliography{main}  

\end{document}